\pdfoutput=1

\documentclass[11pt]{article}

\usepackage[final]{acl}

\usepackage{times}
\usepackage{latexsym}

\usepackage[T1]{fontenc}

\usepackage[utf8]{inputenc}

\usepackage{microtype}

\usepackage{inconsolata}

\usepackage{graphicx}
\usepackage{multirow}
\usepackage{amsmath}
\usepackage{makecell}
\usepackage{graphicx}
\usepackage{placeins}
\usepackage{float}
\usepackage{placeins}
\usepackage{booktabs}
\usepackage{tablefootnote}

\title{PathoHR: Hierarchical Reasoning for Vision-Language Models in Pathology}

\author{
  Yating Huang\textsuperscript{1}\thanks{Equal contribution.}, 
  Ziyan Huang\textsuperscript{2}\footnotemark[1], 
  Lintao Xiang\textsuperscript{1}, 
  Qijun Yang\textsuperscript{1}, 
  Hujun Yin\textsuperscript{1} \\
  \textsuperscript{1}University of Manchester \\
  \textsuperscript{2}South China University of Technology \\
  \texttt{\{yating.huang, hujun.yin\}@manchester.ac.uk}, 
\texttt{bonnie.ziyan.huang@gmail.com}
}

\begin{document}
\maketitle
\begin{abstract}
Accurate analysis of pathological images is essential for automated tumor diagnosis but remains challenging due to high structural similarity and subtle morphological variations in tissue images. Current vision-language (VL) models often struggle to capture the complex reasoning required for interpreting structured pathological reports. To address these limitations, we propose PathoHR-Bench, a novel benchmark designed to evaluate VL models’ abilities in hierarchical semantic understanding and compositional reasoning within the pathology domain. Results of this benchmark reveal that existing VL models fail to effectively model intricate cross-modal relationships, hence limiting their applicability in clinical setting. To overcome this, we further introduce a pathology-specific VL training scheme that generates enhanced and perturbed samples for multimodal contrastive learning. Experimental evaluations demonstrate that our approach achieves state-of-the-art performance on PathoHR-Bench and six additional pathology datasets, highlighting its effectiveness in fine-grained pathology representation.
\end{abstract}

\section{Introduction}
\label{sec:1}
Recently, vision-language (VL) models have gained substantial advances \citep{goel2022cyclip,li2022blip}, fostering an integration of computer vision and natural language processing across a wide range of application domains \citep{wei2024vary,das2024exams}. In medical image analysis, contrastive learning-based approaches have been extensively employed to align large-scale medical images with corresponding diagnostic reports, thereby enabling zero-shot classification and grading without additional fine-tuning \citep{xie2024pairaug,phan2024decomposing}. This paradigm markedly reduces the reliance on high-quality annotated data and enhances the scalability of medical imaging models. Nevertheless, in contrast to other imaging modalities such as X-ray and MRI, pathological image analysis presents distinct and more formidable challenges due to its higher structural complexity and more subtle visual cues in the images.
\begin{figure}
  \centering
  \includegraphics[width=\linewidth]{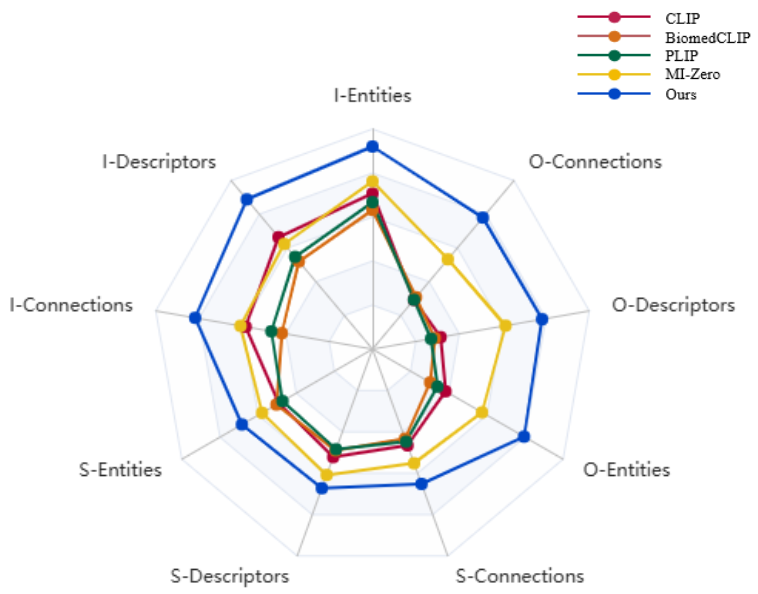}
  \caption {Radar charts for compared models on PathoHR-Bench across multiple compositional reasoning aspects. The axes correspond to three types of perturbations: I (Information Loss), S (Semantic Drift), O (Order Variation), evaluated under three semantic roles: Entities, Descriptors and Connections. Higher values indicate stronger robustness.}
  \label{fig1}
\end{figure}

Pathological biopsy microscopy remain the gold standard for tumor diagnosis, as it capture intricate cellular structures and morphological details \citep{ruiz2017actualizacion}. However, the subtle intra-class variations in images pose significant challenges for disease classification. Existing approaches predominantly rely on task-specific models tailored to individual applications, such as Gleason grading of prostate cancer and breast cancer subtyping \citep{pati2023weakly, bulten2022artificial}. 
In contrast, pathological texts are inherently fine-grained, characterized by specialized medical terminology, precise pathological reasoning, and structured diagnostic logic. These narratives not only describe subtle visual cues but also offer critical insights for diagnostic grading and prognosis prediction.

Although recent studies have explored VL pretraining and zero-shot transfer in pathology~\citep{javed2024cplip, lu2023harnessing, lu2023towards}, most existing models still treat pathological texts as a “bag-of-words,” ignoring their hierarchical structure, syntactic dependencies, and diagnostic reasoning logic. Previous research~\citep{yuksekgonul2022and} reveals that standard VL benchmarks often inadequately evaluate a models’ capacity for compositional and structural semantic understanding. This shortcoming is particularly critical in pathology, where diagnostic texts require more than surface-level pattern matching. Accurate and interpretable decision-making in pathology requires effective modeling of hierarchical clinical language in conjunction with visual and morphological cues. Current models and evaluation methods face several limitations: 

\noindent\textbf{i). Limited hierarchical structure awareness:} Pathological tests typically follow a structured diagnostic pattern \citep{bera2019artificial}, (e.g., lesion region + cellular morphology + symptoms identification + diagnostic conclusion.). However, most VL models treat texts as unordered token sets, lacking the capacity to model multi-level semantic dependencies or structured reasoning patterns. Although these models may perform adequately on retrieval or classification tasks, their disregard for compositional structure undermines their interpretability and alignment with clinical diagnostic logic. 

\noindent\textbf{ii). Limited compositional reasoning capability:} Pathology texts contain complex reasoning that integrates anatomical regions, cellular structures, and molecular markers, often within intricate spatial and functional contexts. Bag-of-words-based models would fail to capture such contextual and inferential relationships, thereby hindering accurate and nuanced diagnostic decision-making.

\begin{figure*}
  \centering
  \includegraphics[width=\linewidth]{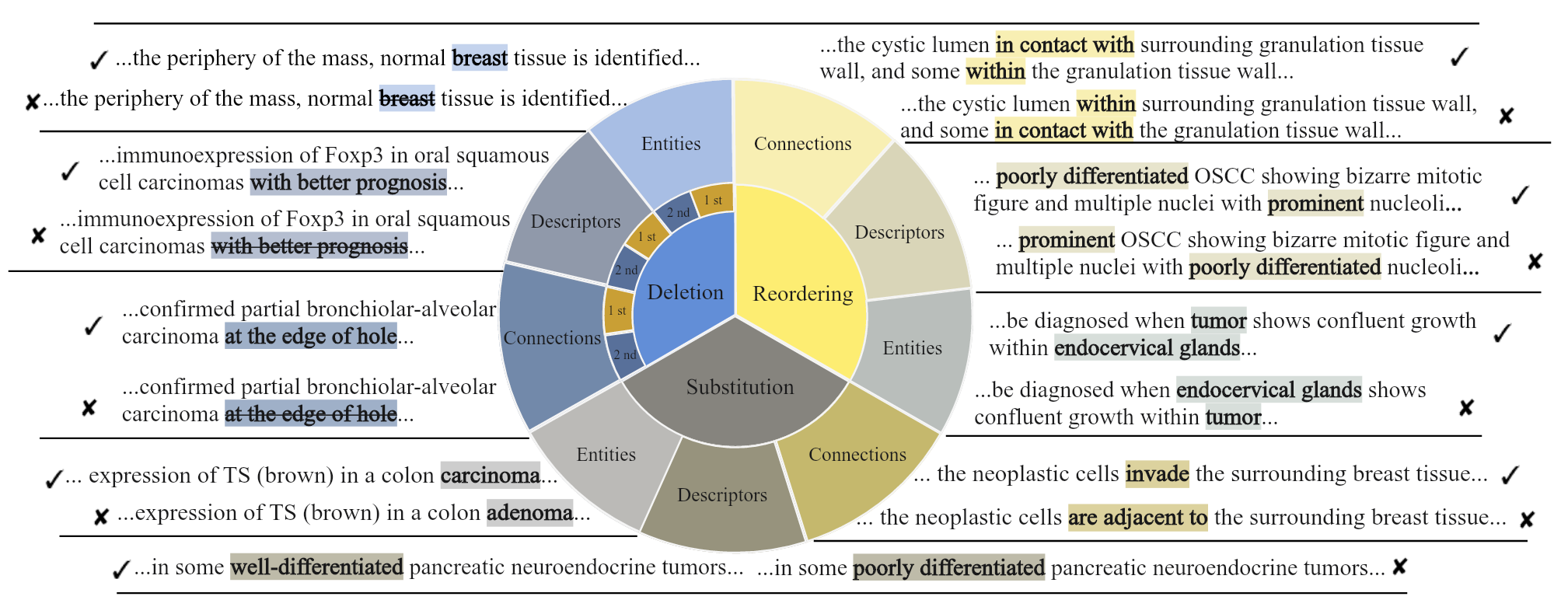}
  \caption {Cross-dimensional taxonomy in PathoHR-Bench: Text perturbation levels and semantic role levels.}
  \label{fig2}
\end{figure*}

Based on the above analysis, the main contributions of this work are summarized as follows:
\begin{itemize}
\item We introduce PathoHR-Bench, a novel benchmark designed to evaluate the hierarchical structure awareness and compositional reasoning capabilities of pathological VL models. Unlike existing evaluation protocols that focus on application performance, PathoHR-Bench offers a systematic assessment of model robustness under three core perturbation types that reflect key challenges in pathology text comprehension. To better reflect the structured nature of diagnostic narratives, the benchmark further categorizes evaluations according to distinct semantic roles commonly observed in pathology reports.

\item We subsequently propose a data-driven VL training scheme tailored to the pathology domain, aimed at more effectively leveraging existing pathological VL datasets. The approach generates both enhanced and perturbed samples across textual and visual modalities, and employs cross-modal contrastive learning to improve the model’s ability to capture fine-grained semantic alignments and diagnostic reasoning cues.
\end{itemize}
As illustrated in Figure~\ref{fig1}, the proposed training scheme yields consistent performance gains across all perturbation types and hierarchical components on the PathoHR-Bench, highlighting its effectiveness in capturing multi-level semantic dependencies and inferring complex pathological relationships. Moreover, in standard zero-shot diagnostic classification tasks, we observe that improved structural understanding and compositional reasoning directly enhance diagnostic accuracy, underscoring the critical importance of these capabilities for real-world clinical deployment. The full benchmark suite and source code will be publicly released upon acceptance.
\section{Motivations}
\label{sec:2}
The lack of hierarchical reasoning capabilities significantly limits the reliability and clinical interpretability of existing VL models in pathological analysis. A further limitation lies in current evaluation practices, which predominantly rely on standard retrieval or classification metrics and offer limited insight into models’ structural understanding or reasoning ability. While recent studies in the natural image domain have proposed benchmarks for evaluating compositional reasoning in VL models \citep{zhao2022vl, diwan2022winoground}, these approaches are not directly applicable to pathology due to fundamental domain differences in textual structure and semantics. Descriptions in natural image datasets are typically open-domain and emphasize visually salient features. 
In contrast, pathological texts are “closed-domain,” relying on highly specialized and standardized vocabularies (e.g., MeSH, ICD-O) and structured diagnostic patterns. Unlike open-domain free text, these reports follow controlled terminology and precise reasoning logic, resulting in much tighter semantic relationships and long-range inferential chains.

For instance, in describing a gastric cancer pathological image: “Disordered glandular arrangement with gland fusion and lumen disappearance, consistent with poorly differentiated adenocarcinoma.” Here, the reasoning between “gland fusion” and “poorly differentiated” determines the malignancy grade of the lesion \citep{grillo2020neoplastic}, making this logical relationship crucial for both model learning and final diagnostic decision-making.
In addition, pathological texts often contain a significant amount of “implicit reasoning”. This type of inference typically requires integration with external medical knowledge bases \citep{jha2017augmenting} rather than solely relying on visual features for text generation. These characteristics underscore the need for a domain-specific benchmark tailored to pathology. In this context, PathoHR-Bench is designed to fill this gap by providing a structured and diagnostic-relevant framework for evaluating the compositional and inferential reasoning capabilities of VL models in the pathological domain.

\begin{figure*}
  \centering
  \includegraphics[width=0.95\linewidth]{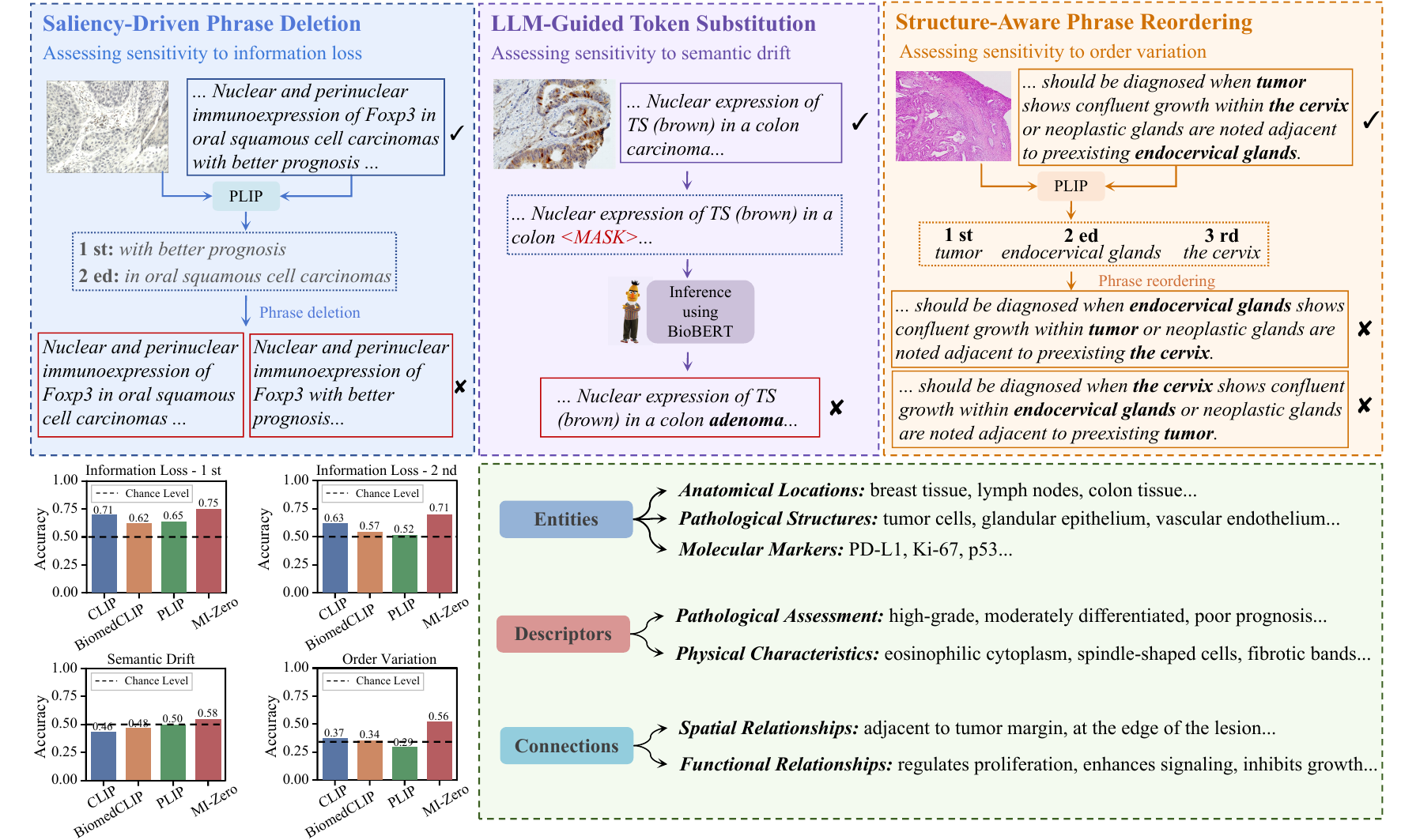}
  \caption {Overview of proposed PathoHR-Bench, comprising three sensitivity tests (top row) with performances of existing VL models (bottom left). Bottom right shows further semantic perturbation levels.}
  \label{fig3}
\end{figure*}
\section{Pathology Hierarchical Reasoning Benchmark (PathoHR-Bench)}
\label{sec:3}
Inspired by VL-CheckList \citep{zhao2022explainable} and ARO \citep{yuksekgonul2023when} that utilize generated adversarial samples and image-text matching as primary evaluation objectives, we propose a novel pathology VL benchmark, termed as PathoHR-Bench, which is designed to systematically assess pathology-related VL models on hierarchical structure awareness and compositional reasoning capabilities. As shown in Figure~\ref{fig2}, it adopts a cross-dimensional structure, enabling a comprehensive assessment of VL models across two independent yet interrelated dimensions: text perturbation and semantic role. This design allows for an in-depth analysis of model robustness, fine-grained comprehension, and compositional reasoning when processing pathological VL data. By identifying model limitations more effectively, PathoHR-Bench can serve as a valuable resource for advancing VL model development and optimization for pathology applications. 
Details of the PathoHR-Bench are illustrated in Figure~\ref{fig3}.

At the text perturbation level, we have designed three core tasks to simulate the challenges encountered in real-world pathology texts, including information loss, semantic drift, and order variation:

\noindent\textbf{i). Assessing sensitivity to information loss:}
We employ saliency-driven phrase deletion by leveraging the Pathology Language Image Pretraining (PLIP) \citep{huang2023visual} model to compute the similarity between each phrase and the corresponding image. The two most salient pathological terms are removed to create adversarial text variants, simulating the model's reliance on critical textual cues and assessing its behavior under different deletion orders. It reflects how the model allocates information between visual and textual inputs while evaluating its robustness to local information losses.

\noindent\textbf{ii). Assessing sensitivity to semantic drift:}
We employ LLM-guided token substitution, where a randomly selected word in each text is masked and then predicted using a pretrained medical BERT model (BioBERT) \citep{kenton2019bert,lee2020biobert}. The generated adversarial samples test the model’s robustness to semantic drifts and fine-grained semantic comprehension. It evaluates whether the VL models can accurately differentiate between subtly different but diagnostically significant textual descriptions.

\noindent\textbf{iii). Assessing sensitivity to order variation:}
We employ structure-aware phrase reordering to examine the model’s sensitivity to order variation. Specifically, we use PLIP \citep{huang2023visual} to identify the top three phrases with the highest image-text similarity and cyclically reorder them within the sentence to generate two adversarial variants. This strategy assesses whether the VL model responds to sentence structure changes, offering insights into its structural awareness.

As shown in the bottom-left of Figure~\ref{fig3}, current pathological VL models perform poorly on perturbation-level evaluation tasks, often at or below random chance which highlighting a lack of structural awareness and fine-grained compositional reasoning abilities. Although VL models can leverage shortcut strategies~\citep{geirhos2020shortcut} during contrastive pretraining to excel at coarse-grained classification and matching, such heuristics are insufficient for pathology-specific tasks. The image-text relationships in pathology are more complex than in general vision tasks, requiring hierarchical semantic understanding and the modeling of nuanced diagnostic relations.

To further investigate these limitations, we extend the perturbation-level evaluation by introducing three major semantic roles derived from the structural patterns of pathological texts.

\noindent\textbf{i). Entities:} Including anatomical locations, pathological structures, and molecular markers, which form the foundation of pathology reports. VL models must accurately recognize these entities to achieve precise image-text alignment.

\noindent\textbf{ii). Descriptors:} Including terms that modify pathological entities, such as diagnostic assessment terms and physical characteristics. These descriptors are crucial for precise disease classification.

\noindent\textbf{iii). Connections:}
This category consists of textual elements describing spatial relationships and functional interactions between entities. Such information facilitates complex pathological reasoning and disease progression analysis. 

Examples of these semantic role classifications are illustrated in Figure~\ref{fig3}. By combining the semantic role hierarchy with text perturbation tasks, we construct a systematic, cross-dimensional evaluation framework. The key advantage of this framework is its ability to assess VL models not only for robustness under various text transformations but also for their understanding of different pathology-related semantic concepts. This allows for a more precise identification of model weaknesses and developing effective optimisation strategies. A detailed evaluation of various models on PathoHR-Bench is presented in Section~\ref{sec:5}.
\begin{figure*}
  \centering
  \includegraphics[width=\linewidth]{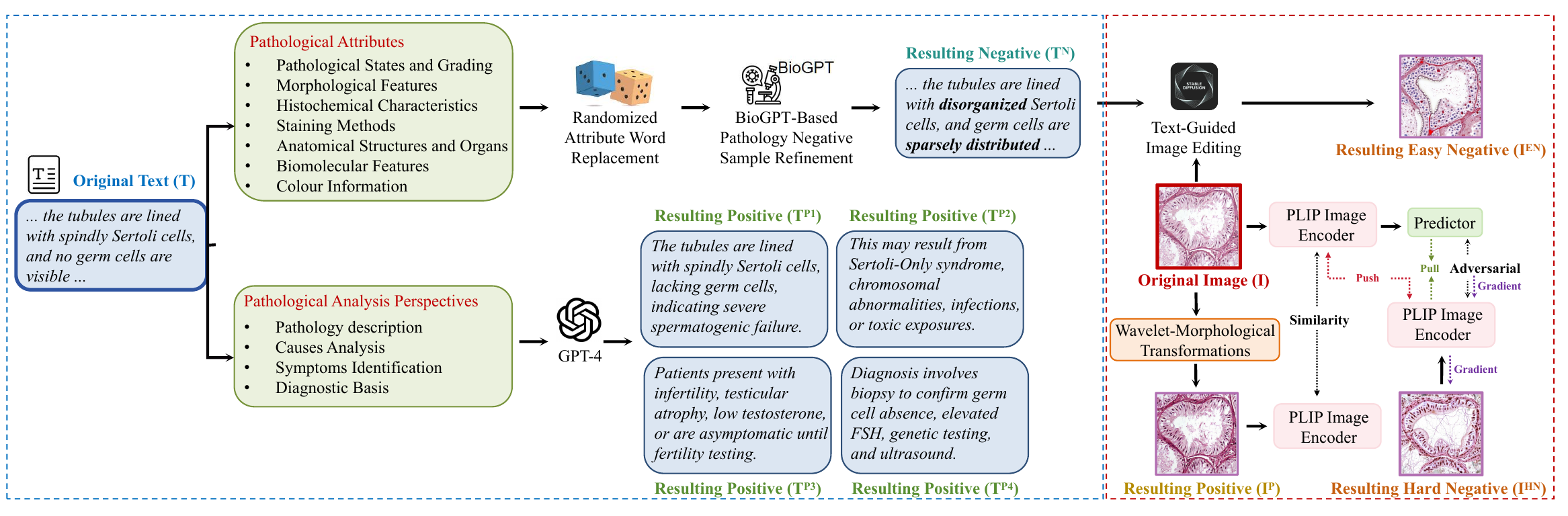}
  \caption {Structured textual and visual data manipulation for pathological VL training.}
  \label{fig4}
\end{figure*}
\section{Pathology Hierarchical Reasoning Training Scheme}
\label{sec:4}
We propose a data-driven training scheme for pathological VL model, shown in Figure~\ref{fig4}. It consists of four branches for generating positive and negative samples for both text and imagery. The following subsections discuss these four branches and the losses incorporated into VL contrastive learning.

\subsection{Pathology-guided textual perturbation}
\label{sec:4.1}
A simple but effective approach of negative text generation is to apply a collection of pre-defined linguistic rules to match and replace words associated with specific entity types or patterns \citep{doveh2023teaching}. We pre-defined seven pathological attribute dimensions (pathological states and grading, morphological features, histochemical characteristics, staining methods, anatomical structures and organs, biomolecular features, color information) via PubMed \citep{namata:mlg12} and MeSH \citep{lipscomb2000medical}, followed by random substitution at the level of pathological attributes to generate controlled perturbations. 

To account for the specificity of pathological language, we further introduce a refinement stage using BioGPT \citep{luo2022biogpt}, which ensures that the generated negative texts retain pathological plausibility and clinical relevance. This pathology-guided textual perturbation applies controlled semantic transformations and plausibility correction, enabling VL models to learn subtle yet critical differences in pathological semantics.

\subsection{Hierarchical diagnostic reasoning-based text expansion}
\label{sec:4.2}
This branch provides multi-level structured diagnostic insights, guiding the model to engage in hierarchical compositional reasoning and allowing it to grasp diagnostic logic rather than relying on superficial text-image correlations. Traditional VL models tend to rely heavily on shallow semantic matching, ignoring the reasoning process and hierarchical context of diagnosis. To address this, we employed GPT-4 \citep{achiam2023gpt} to generate structured positive texts across four pathological analysis perspectives: pathological description, causes analysis, symptoms identification, and diagnostic basis. This approach enables VL models to learn from multi-perspective and multi-level diagnostic information, enhancing their compositional reasoning in medical knowledge. By incorporating hierarchical diagnostic reasoning, the model shifts from data-driven medical text generation to more explainable diagnostic reasoning, improving its contextual reasoning capabilities.

\subsection{Dual-constraint negative image mining}
\label{sec:4.3}
To enhance the fine-grained structural reasoning ability of VL models in pathology, we propose a dual-constraint negative sample mining strategy. This approach generates negative images under two complementary mechanisms: semantic inconsistency and distributional ambiguity, yielding both easy and hard negatives. 

\noindent{\textbf{i. Text-guided visual editing:}
To simulate explicit semantic inconsistencies, we employ text-guided image editing using Stable Diffusion~\citep{rombach2022high} to generate easy negative images conditioned on deliberately corrupted diagnostic texts. These generated samples are not expected to conform to natural image distributions. Instead, they serve as visual exaggerations of erroneous semantics, providing strongly misaligned image-text pairs. The goal is not to simulate realistic pathology, but to reinforce the model’s ability to identify and reject pathology-irrelevant or structurally invalid cues. This mechanism enhances fine-grained semantic-to-structural reasoning by anchoring the model's attention to diagnostically critical patterns.

\noindent{\textbf{ii. Adversarial distribution-aware perturbation:} In parallel, we introduce an adversarial distribution-aware sampling strategy that generates hard negatives without relying on textual input. Operating purely in the visual domain, this module seeks perturbations near the real distribution of pathological images that would cause the most confusion to the model. We consider the worst-case scenario~\cite{qiao2020learning} around the original distribution $M_{0}$:
\begin{equation}
\min_{\theta} \sup_{M_{N}: D\left ( M_{N},M_{0}  \right )\le m  } E_{M_{N}}\left [ L\left ( \theta , I  \right)  \right ]  
\label{eq00}
\end{equation}

\noindent where $\theta$ denotes the pretrained model weights, $I$ is the original image, and $L$ represents the task-specific loss. $D$ denotes the distance metric, and $m$ is the maximum distributional variability between $M_0$ and the generated distribution $M_N$.

The solution to worst-case problem aims to generate negative samples that remain visually and statistically plausible within the pathology domain, while being maximally separated from the original distribution under the constraint. These samples serve as structurally valid but distributionally challenging examples, exposing the model to difficult decision-boundary cases without deviating from the natural appearance of real pathological images. Detailed formulation and optimization procedures are provided in the Appendix~\ref{sec:appendix0}.

\subsection{Wavelet-morphology-guided consistency refinement}
\label{sec:4.4}
This branch enhances the multi-scale representation of pathology images while preserving semantic consistency of tissue structures. A wavelet transform~\citep{othman2020applications} decomposes each image into frequency components, with high-frequency bands capturing microstructural details. Morphological operations are then applied to these components: Top-Hat enhances bright structures (e.g., nuclei, fibers), Black-Hat emphasizes low-density regions (e.g., necrosis, inflammation), and Morphological Gradient highlights cellular boundaries. The enhanced image is reconstructed via inverse wavelet transform. To ensure semantic alignment with the original image, we apply a consistency constraint based on feature similarity computed using a pretrained PLIP model. A similarity threshold dynamically adjusts transformation parameters, generating structurally enriched positive samples that preserve pathological integrity.

\subsection{Loss Function}
\label{sec:4.5}
In our validation experiments, the dual-encoder vision-language models (e.g., CLIP \citep{radford2021learning}, PLIP \citep{huang2023visual}) admit text-image pairs $\left ( T,I \right )$ through a text encoder $f_{T}\left ( T \right )$ and an image encoder $f_{I}\left ( I \right )$. The text-image similarity score is then computed as:
\begin{equation}
\resizebox{0.75\hsize}{!}{$
S\left ( T,I \right ) =exp\left ( \frac{\alpha f_{T}\left ( T \right )^{T} f_{I}\left ( I \right )  }{\left \| f_{T}\left ( T \right ) \right \|^{2}\left \| f_{I}\left ( I \right ) \right \|^{2}  }  \right )
$}
\label{eq0}
\end{equation}
where $\alpha$ is a learned temperature parameter.

\noindent\textbf{Contrastive Loss.} Similar to most contemporary VL models, we employ the contrastive CLIP-loss as one of the losses for each batch $B=\left \{ \left ( T_{i}, I_{i}, \right )  \right \} $:
\begin{equation}
\resizebox{0.95\hsize}{!}{$
L_{con} = \sum_i \log \left( \frac{S(T_i, I_i)}{\sum_j S(T_i, I_j)} \right) + \log \left( \frac{S(T_i, I_i)}{\sum_k S(T_k, I_i)} \right)
$}
\label{eq1}
\end{equation}
\noindent\textbf{Negative Loss.} For the negative text $T_i^N$ generated in Section~\ref{sec:4.1}, we focus on similarity difference between the original text $T_i$ and the generated $T_i^N$ with respect to their corresponding original image $I_i$, leading to a negative text loss:
\begin{equation}
\resizebox{0.8\hsize}{!}{$
L_{neg}^{text}=\sum_i-\log \left(\frac{S\left(T_i, I_i\right)}{S\left(T_i, I_i\right)+S\left(T_i^N, I_i\right)}\right)
$}
\label{eq2}
\end{equation}

Similarly, for the negative images $I_i^{EN}$ and $I_i^{HN}$ generated in Section~\ref{sec:4.3}, we obtain negative image loss $L_{neg}^{img_E}$ and $L_{neg}^{img_N}$ similar to Equation~\ref{eq2}. Then the total negative loss $L_{neg}$ is obtained:
\begin{equation}
L_{neg}=L_{neg}^{text}+L_{neg}^{img_E}+L_{neg}^{img_N}
\label{eq3}
\end{equation}

\noindent\textbf{Positive Loss.}
For the positive text, to ensure that the four hierarchical levels of positive text samples $\left ( T_{i}^{P^{k} },k=1,2,3,4  \right )$ generated in Section~\ref{sec:4.2} are closely aligned in the feature space and matching with the original text $T_{i}$, we compute the cosine similarity between their text embeddings using $S\left ( T_{1}, T_{2} \right )  $ and define the text-text positive loss as:
\begin{equation}
\resizebox{0.8\hsize}{!}{$
L_{pos}^{text_{T}}=\sum_i-\log\left(\frac{ {\textstyle \sum_{k=1}^{4}S\left ( T_{i}^{P_{k} },T_{i}\right ) }}{\sum_j \sum_{k=1}^4 S\left(T_i^{P_k}, T_j\right)}\right)
$}
\label{eq4}
\end{equation}

To ensure that all four hierarchical levels of positive text samples align with the original image $I_i$ and preserve cross-modal consistency, we introduce a text-image positive loss:

\begin{equation}
\resizebox{0.8\hsize}{!}{$
L_{pos}^{text_{I}}=\sum_i-\log\left(\frac{ {\textstyle \sum_{k=1}^{4}S\left ( T_{i}^{P_{k} },I_{i}\right ) }}{\sum_j \sum_{k=1}^4 S\left(T_i^{P_k}, I_j\right)}\right)
$}
\label{eq5}
\end{equation}

\begin{table*}[t]
\centering
\resizebox{\textwidth}{!}{
\begin{tabular}{l|ccc|ccc|ccc|ccc}
\toprule
\multirow{2}{*}{Model} & \multicolumn{3}{c|}{Information Loss-1st} & \multicolumn{3}{c|}{Information Loss-2nd} & \multicolumn{3}{c|}{Semantic Drift} & \multicolumn{3}{c}{Order Variation} \\
\cmidrule(lr){2-13}
& Entities & Descriptors & Connections & Entities & Descriptors & Connections & Entities & Descriptors & Connections & Entities & Descriptors & Connections \\
\midrule
CLIP         & 0.7072 & 0.6643 & 0.5851 & 0.6431 & \underline{0.6129} & 0.5237 & 0.4915 & 0.5220 & 0.4655 & 0.3831 & 0.3026 & 0.3409 \\
BiomedCLIP   & 0.6325 & 0.5201 & 0.4138 & 0.5924 & 0.5088 & 0.4152 & 0.5037 & 0.4904 & 0.4326 & 0.3142 & 0.2874 & 0.2703 \\
PLIP         & 0.6679 & 0.5472 & 0.4664 & 0.5210 & 0.4936 & 0.4427 & 0.4720 & 0.4843 & 0.4473 & 0.2926 & 0.3079 & 0.2931 \\
CONCH        & 0.6531 & 0.5496 & 0.4389 & 0.6031 & 0.5142 & 0.4195 & 0.5028 & 0.5288 & 0.4736 & 0.3574 & 0.4397 & 0.4052 \\
QuiltNet     & 0.7354 & \underline{0.6992} & \underline{0.7158} & 0.6985 & 0.6079 & \underline{0.6423} & 0.5529 & 0.5706 & 0.5230 & 0.4572 & 0.5831 & 0.4590 \\
MI-Zero      & \underline{0.7623} & 0.6236 & 0.6089 & \underline{0.7114} & 0.6033 & 0.5796 & \underline{0.5789} & \underline{0.6083} & \underline{0.5519} & \underline{0.5748} & \underline{0.6139} & \underline{0.5327} \\
\midrule
\textbf{Ours} & \textbf{0.9134} & \textbf{0.8780} & \textbf{0.8205} & \textbf{0.8901} & \textbf{0.8427} & \textbf{0.7945} & \textbf{0.6853} & \textbf{0.6744} & \textbf{0.6520} & \textbf{0.7932} & \textbf{0.7819} & \textbf{0.7813} \\
\bottomrule
\end{tabular}
}
\caption{\label{1}Performance comparison across different compositional reasoning aspects on PathoHR-Bench. Best results are highlighted in \textbf{bold} and second best in \underline{underline}.}
\end{table*}
\begin{table*}[htbp]
  \centering
  \fontsize{8}{9.5}\selectfont
  \resizebox{\textwidth}{!}{
  \begin{tabular}{l|cc|cc|cc|cc|cc|cc}
    \toprule
    \multirow{2}{*}{\textbf{Model}} & \multicolumn{2}{c|}{\textbf{CRC100K}} & \multicolumn{2}{c|}{\textbf{UHU}} & \multicolumn{2}{c|}{\textbf{PanNuke}} &
    \multicolumn{2}{c|}{\textbf{DigestPath}}&
    \multicolumn{2}{c|}{\textbf{TCGA-BRCA}}&
    \multicolumn{2}{c}{\textbf{TCGA-RCC}}\\
    \cmidrule(lr){2-13}
    & \textbf{Acc} & \textbf{F1} & \textbf{Acc} & \textbf{F1} & \textbf{Acc} & \textbf{F1} & \textbf{Acc} & \textbf{F1}& \textbf{Acc} & \textbf{F1}& \textbf{Acc} & \textbf{F1}\\
    \midrule
  CLIP & 0.2593 & 0.1971 & 0.3349 & 0.1836 & 0.3220 & 0.3517 & 0.2056 & 0.1432& 0.5021 & 0.3490 & 0.3348 & 0.1956\\
  BiomedCLIP & 0.4461 & 0.3549 & 0.3538 & 0.2281 & 0.5632 & 0.5926 & 0.6149 & 0.5974 & 0.5164 & 0.4209 & 0.6533 & 0.6273\\
  PLIP & 0.5361 & 0.4603 & \underline{0.3735} & 0.2538 & 0.6275 & 0.6438 & 0.7955 & 0.8031& 0.4508 & 0.3502 & 0.6987 & 0.7055\\
  CONCH & 0.5482 & 0.5219 & 0.3574 & 0.2306 & 0.5074 & 0.5128 & \underline{0.8386} & \underline{0.8470} & 0.6327 & 0.6152 & 0.7899 & 0.7740\\
  QuiltNet & 0.5097 & 0.4390 & 0.3413 & 0.2451 & 0.6377 & 0.5932 & 0.8059 & 0.7814 & 0.7235 & 0.6943 & \underline{0.8053} & \underline{0.7905}\\
  MI-Zero & \underline{0.5637} & \underline{0.5626} & 0.3492 & \underline{0.2870} & \underline{0.6527} & \underline{0.6779} & 0.8253 & 0.8214 & \underline{0.7729} & \underline{0.7006} & 0.7986 & 0.7260\\
  \midrule
  \textbf{Ours} & \textbf{0.6985} & \textbf{0.6841} & \textbf{0.4937} & \textbf{0.4612} & \textbf{0.7386} & \textbf{0.7609} & \textbf{0.8517} & \textbf{0.8638} & \textbf{0.8129} & \textbf{0.7746} & \textbf{0.8327} & \textbf{0.7996} \\
  \bottomrule
  \end{tabular}
  }
\caption{\label{2}
  Zero-shot classification performance using a single prompt per class, reported as balanced accuracy and weighted F1 across \textbf{six} datasets. Best and second results are highlighted with \textbf{bold} and \underline{underline}.}
\end{table*}

For the positive image loss, we omit the image-image positive term, as the generation process described in Section~\ref{sec:4.4} ensures feature consistency with the original image. Thus, our goal is to ensure that the positive images $I_i^P$ remain closely aligned with their corresponding text descriptions:

\begin{equation}
\resizebox{0.75\hsize}{!}{$
L_{pos}^{img}=\sum_i-\log \left(\frac{S\left(T_i, I_i^P\right)}{\sum_j S\left(T_j, I_i^P\right)}\right)
$}
\label{eq6}
\end{equation}

The total positive loss $L_{pos}$ is computed as:
\begin{equation}
L_{pos} = L_{pos}^{text_{T}}+L_{pos}^{text_{I}}+L_{pos}^{img}
\label{eq10}
\end{equation}

Finally, the full fine-tuning loss of our proposed method can be written as:
\begin{equation}
L = L_{con}+\alpha \cdot L_{neg}+\beta  \cdot L_{pos}
\label{eq11}
\end{equation}

\section{Experiments}
\label{sec:5}
\subsection{Datasets}
The ARCH dataset \citep{gamper2021multiple} is the only widely available pathology-specific paired image–text dataset, comprising 8,617 pairs extracted from pathology textbooks and PubMed research articles. To avoid overlap between training and evaluation, we carefully split ARCH into two disjoint subsets: textbook-derived samples were used to construct PathoHR-Bench, while PubMed-derived samples were reserved for training. Building on this resource, PathoHR-Bench was created by systematically applying three types of perturbations across three semantic roles. This expansion yielded a total of 77,553 image–text pairs for evaluation. The textbook portion provides more structured and hierarchical narratives, making it especially suitable for robust assessment of reasoning capabilities, while the PubMed portion ensures fair training for all baseline models.

We compared the performance of our proposed method with current pathological VL models, including baseline CLIP \citep{radford2021learning}, PLIP \citep{huang2023visual}, BiomedCLIP \citep{zhang2023biomedclip}, CONCH \citep{lu2024visual}, QuiltNet \citep{ikezogwo2023quilt}, and MI-Zero \citep{lu2023visual} on PathoHR-Bench.
Inspired by the fair comparison strategy in CPLIP \citep{javed2024cplip}, we fine-tuned all baseline models on the ARCH dataset before evaluation, with the exception of CLIP. Notably, while PLIP was originally pre-trained on PubMed captions and various biomedical image–text pairs, it had not been fine-tuned on ARCH in its released version. To ensure consistent data distribution and minimize domain bias, we additionally fine-tuned PLIP and the other baselines on ARCH using their original loss objectives, without introducing our perturbation-based augmentations. CLIP remained a purely zero-shot baseline without fine-tuning on ARCH, serving as a general-purpose reference point. All models were then evaluated under the same testing splits and inference prompts for fair comparison.

For the zero-shot task, we utilized six publicly available datasets covering a range of cancer types, including four datasets at the patch level and two at the whole-slide level. To ensure fairness and consistency across all methods, we adopted a single prompt per class for evaluation, rather than using prompt ensembling. Detailed dataset descriptions are provided in the Appendix~\ref{sec:appendix1}, and implementation details can be found in the Appendix~\ref{sec:appendix11}. To disentangle the effect of ARCH-specific fine-tuning from the inherent representation ability of existing models under our unified single-prompt protocol, we additionally report the performance of BiomedCLIP and PLIP \emph{without} any fine-tuning on ARCH in Appendix~\ref{sec:appendixd}. 

\subsection{Comparison of existing VL models}
To comprehensively evaluate the current pathological VL models, we conducted a detailed assessment using the benchmark proposed in Section~\ref{sec:3}. Results are shown in Table~\ref{1}. The results reveal an unexpected fact that some existing VL models that were trained specifically on pathological datasets performed worse in structural awareness and compositional reasoning tests for pathological texts compared to the CLIP baseline trained on natural images. 

This discrepancy may be attributed to the limited scale of pathological datasets and the rich and diverse open-world concepts in natural image-text pairs, which may inherently promote compositional reasoning capability. However, CLIP performed poorly in pathology-specific tasks, demonstrated by near-zero effectiveness as shown in Table~\ref{2}, as expected. These results highlight a fundamental limitation of current pathological VL models that domain adaptation may help improve zero-shot performance in pathological classification tasks, it remains insufficient to address the complexity of pathological reasoning.

In contrast, our proposed method effectively leverages limited pathology-specific data to enable fine-grained compositional reasoning while maintaining strong performance on conventional classification tasks.
As shown in Tables~\ref{1} and~\ref{2}, the model achieves consistent improvements across nine textual perturbation settings and six zero-shot classification benchmarks. Notably, substantial gains are observed on CRC100K and UHU (with fine-grained cancer subtypes), while more moderate gains are seen on PanNuke and DigestPath (with coarser tumor/normal classification). These results further support PathoHR-Bench as a robust benchmark for evaluating a VL model’s ability to capture fine-grained pathological semantics. The case studies provided in the appendix~\ref{sec:appendix2} further illustrate the model's performance on fine-grained diagnosis.

\begin{table*}[htbp]
  \centering
  \fontsize{8}{9.5}\selectfont
  \resizebox{\textwidth}{!}{
  \begin{tabular}{l|ccc|ccc|ccc|c}
    \toprule
    \multirow{2}{*}{\textbf{Model}} & \multicolumn{3}{c|}{\textbf{Information Loss}} & \multicolumn{3}{c|}{\textbf{Semantic Drift}} & \multicolumn{3}{c|}{\textbf{Order Variation}} & \multirow{2}{*}{\makecell{\textbf{6 Zero-shot} \\ \textbf{Tasks Average}}}\\
    \cmidrule(lr){2-10}
    & \textbf{Ent.} & \textbf{Desc.} & \textbf{Conn.} & \textbf{Ent.} & \textbf{Desc.} & \textbf{Conn.} &  \textbf{Ent.} & \textbf{Desc.} & \textbf{Conn.} & \\
    \midrule
    CLIP & 0.6752 & 0.6386 & 0.5544 & 0.4915 & 0.5220 & 0.4655 & 0.3831 & 0.3026 & 0.3409 & 0.3725 \\
    PLIP & 0.5945 & 0.5204 & 0.4546 & 0.4720 & 0.4843 & 0.4473 & 0.2926 & 0.3079 & 0.2931 & 0.5246 \\
    \midrule
    Ours Text Neg & 0.8896 & 0.8637  &   0.7833 & \textbf{0.7032} & 0.6651 & \underline{0.6509} & 0.7863 & 0.7640 & 0.7698 & 0.5892\\
    Ours Text+Img Neg & 0.9005 & 0.8720 & 0.7926 & 0.6924 & 0.6704 & 0.6395 & \underline{0.7914} & 0.7729 & 0.7520 & 0.5804 \\
    Ours Text Pos & 0.8032 & 0.8154 & 0.6941 & 0.6538 & 0.6370 & 0.5786 & 0.5249 & 0.5462 & 0.5033 & 0.6497 \\
    Ours Text+Img Pos & 0.7976 & 0.8203 & 0.6859 & 0.6249 & 0.6459 & 0.5842& 0.5087 & 0.4576 & 0.4725 & 0.6718\\
    \midrule
     Ours w/o Img Neg$_{Easy}$ & 0.8529	& 0.8342	& 0.7803&	0.6625&	0.6431&	0.6319&	0.7635&	0.7796&	0.7446&	\underline{0.6728}\\
    
     Ours w/o Img Neg$_{Hard}$ & 0.8973 & 0.8675 & \underline{0.8018} & \underline{0.6949} & \underline{0.6724} & 0.6507 & 0.7911 & \underline{0.7804} & \underline{0.7801} & 0.6537\\
    
    Ours w/o Text Pos&	0.8742&	0.8631&	0.7746&	0.6731&	0.6522&	0.6425&	0.7806&	0.7649&	0.7537&	0.6293\\
   
    Ours w/o Img Pos&	\underline{0.9022}&	\textbf{0.8791} &	0.7854&	0.6780&	0.6638&	0.6409&	0.7824&	0.7725&	0.7679&	0.6358\\
    \midrule
    \textbf{Ours Combined} & \textbf{0.9018} & \underline{0.8604} & \textbf{0.8075} & 0.6853 & \textbf{0.6744} & \textbf{0.6520} & \textbf{0.7932} & \textbf{0.7819} & \textbf{0.7813} & \textbf{0.7380}\\
    \bottomrule
  \end{tabular}
  }
  \caption{\label{3}
    Ablation study on the contribution of different components. Best and second results are highlighted with \textbf{bold} and \underline{underline}.
  }
\end{table*}

\subsection{Ablation studies}
We performed a series of ablation studies to validate the effectiveness of the proposed key components in 
Section~\ref{sec:4}, including negative text (Text Neg) generated via textual perturbation, positive text (Text Pos) derived from diagnostic reasoning-based expansion, easy and hard negative images (Img Neg$_\text{Easy}$ and Img Neg$_\text{Hard}$) obtained through a dual-constraint negative sample mining strategy, and positive images (Img Pos) refined by wavelet-morphology-guided consistency.

Results are presented in the Table~\ref{3}, which summarizes the individual contributions to model performance on PathoHR-Bench, along with average zero-shot accuracy across six pathology datasets. We observe that incorporating either negative text or negative images enhances structural awareness, but alone is insufficient to boost zero-shot performance. As detailed in Appendix~\ref{sec:appendix3}, the generation of negative samples introduces contrastive, parallel, and inclusion relationships; among them, inclusion relationships often lead to semantic confusion due to high similarity, which undermines contrastive learning effectiveness.

In contrast, positive text and images improve classification accuracy but may reduce the model’s sensitivity to structural variations. Our full scheme achieves optimal results by jointly enhancing structural understanding, compositional reasoning, and generalization ability. To further assess the clinical plausibility of the generated samples, we conducted a qualitative evaluation with a certified pathologist. The evaluation protocol and results are presented in Appendix~\ref{sec:appendix4}.
\section{Conclusions}
In this study, we investigated the limitations of current VL models in pathological image analysis and introduced a new benchmark to evaluate their structure awareness and compositional reasoning on pathology-specific text-image pairs. To the best of our knowledge, this is the first work to explicitly focus on compositional reasoning in pathological VL models. By targeting fundamental reasoning capabilities, PathoHR-Bench provides deep insights into current model shortcomings and can foster methodological advancements in pathology-specific VL understanding. In addition, we proposed a data-driven training scheme to enhance the fine-grained learning capacity of existing pathological VL models. Through diverse augmentations and perturbations across four branches, our approach not only improves the model’s fine-grained reasoning ability but also achieves notable gains in zero-shot pathological diagnosis tasks.

\section*{Potential Limitations}
Despite the improvements achieved with PathoHR-Bench and our proposed training scheme, several limitations remain. First, while the proposed benchmark systematically evaluates structural awareness and compositional reasoning, it does not yet incorporate external medical knowledge, which could further enhance model interpretability and diagnostic accuracy. Second, although our negative sample construction was carefully designed to simulate realistic yet diagnostically challenging perturbations with both clinical plausibility and semantic precision, its scalability remains limited. Future work could explore more automated approaches, such as leveraging large language models to simplify perturbation generation and improve adaptability. Third, the perturbation strategies used in the framework, though effective, may not fully capture the diverse and nuanced variations present in real-world pathology reports.

Additionally, our approach to model training is data-driven, meaning that its performance is still constrained by the availability and quality of existing pathology datasets. Fourth, although we conducted an expert evaluation to assess the clinical plausibility of generated samples, it was based on a limited sample set and a single pathologist’s judgment. Broader expert participation, inter-rater agreement analysis, and context-aware evaluations are needed to fully validate the diagnostic reliability and safety of generated content. 

Future work should explore larger-scale multimodal datasets, integrate explicit medical reasoning modules, and refine adaptive augmentation techniques to further improve the robustness and clinical applicability of pathology-focused VL models. 
\section*{Ethics Statement}
The data utilized in our study was sourced from
public repositories, and does not pose any privacy concerns. We are confident that our research
adheres to the ethical standards set forth by ACL.

\bibliography{custom}
\appendix
\renewcommand{\thefigure}{A\arabic{figure}} 
\renewcommand{\thetable}{A\arabic{table}}
\setcounter{figure}{0}
\setcounter{table}{0}

\section{Adversarial distribution-aware perturbation}
\label{sec:appendix0}
This section provides the mathematical formulation and optimization process of the adversarial distribution-aware perturbation branch used for hard negative sample generation in our method.

To expose the model to visually plausible yet structurally challenging examples, we aim to generate hard negative images that deviate from the source domain distribution $M_0$ but still remain within the manifold of valid pathological appearances. We consider the worst-case scenario around the source distribution $M_0$ as Equation~\eqref{eq00}, its solution guarantees good performance against data distributions that are distance $m$ away from $M_{0}$. The Wasserstein distance~\cite{ozair2019wasserstein} is used as the distance metric $D$, which is defined by calculating the minimum transport cost between two distributions in the representation space. Assuming that $X_{N}$ and $X_{0}$ are obtained by sampling from the generated distributions $M_{N}$ and original distribution $M_{0}$, we can measure the distance in the representation space as follows:
\begin{equation}
\resizebox{0.99\hsize}{!}{$
D_{\theta}\left ( M_{N},M_{0}  \right ) = inf_{\gamma\in \Gamma \left ( M_{N},M_{0} \right )}E_{\gamma  }\left [ \left \| f\left ( \theta ,X_{N}  \right )-f\left ( \theta ,X_{0}  \right )  \right \|  \right ]_{2}^{2}
$}
\label{eq01}
\end{equation}
\noindent where $f\left ( \theta ,X_{N}  \right )$ is the encoder of the contrastive learning model, $\gamma$ is the joint probability distribution that satisfies the marginal distribution for $M_{N}$ and $M_{0}$, and $\Gamma$ is the set of all joint probability distributions that satisfy the marginal distribution.

By substituting the formula for the distance metric defined in Equation~\eqref{eq01} into the worst-case objective Equation~\eqref{eq00}, the generated distribution $M_{N}$ that satisfies $D\left ( M_{N},M_{0}  \right )\le m$ achieves maximal divergence in the feature space while preserving pathological plausibility and structural coherence with respect to the source domain. However, directly solving for the supremum over the constrained distributional space is intractable. To address this, we introduce a Lagrangian relaxation with a penalty term $\lambda $ that softly enforces the adversarial constraint. The final formulation for adversarial distribution-aware hard negative generation becomes:
\begin{equation}
\min_{\theta} \sup_{M_{N}} E_{M_{N}}\left [ L\left ( \theta ,I \right ) -\lambda D_{\theta } \left ( M_{N},M_{0}  \right )  \right ]
\label{eq7}
\end{equation}

\section{Datasets}
\label{sec:appendix1}
We used six independent publicly available computational pathology datasets, including four at the patch level and two at the whole-slide level:

\textbf{i). CRC100K} \citep{kather2019predicting}: is a colorectal cancer dataset comprising $224\times 224$ pixel image patches acquired at $0.5 \mu m$ per pixel resolution from 50 patients. It includes nine tissue categories: colorectal adenocarcinoma epithelium, normal mucosa, smooth muscle, lymphocytes, mucus, cancer-associated stroma, adipose tissue, background, and debris. The official split consists of 100,000 training and 7,180 testing images. For zero-shot tile-level classification, we directly evaluated on the testing split without any fine-tuning.

\textbf{ii). UHU} \citep{arvaniti2018automated}: is a prostate cancer dataset comprising five tissue microarrays with a total of 886 tissue cores, each sized at $3,100\times 3,100$ pixels. All slides were scanned at $40\times$ magnification using a NanoZoomer scanner. An experienced pathologist annotated benign (BN) regions and three cancer grades (Gleason grade 3, 4, and 5) with pixel-level segmentation masks. We adopted the official preprocessing pipeline from \citep{arvaniti2018automated}, generating a total of 22,022 image patches ($750\times 750$ pixels) after excluding patches dominated by luminal or unannotated areas. The training set contains 2,076 BN, 6,303 grade 3, 4,541 grade 4, and 2,383 grade 5 patches. The test set includes 127 BN, 1,602 grade 3, 2,121 grade 4, and 387 grade 5 patches. Only the patch-level test set was used for zero-shot evaluation.

\textbf{iii). PanNuke} \citep{gamper2019pannuke}: is a multi-organ dataset designed for nuclei segmentation and classification, encompassing 19 tissue types across various pathological conditions. It contains 4,346 training and 1,888 testing images, each sized at $256\times 256$ pixels. Following the evaluation protocol used in PLIP, we assess zero-shot performance on the testing split for a binary classification task that distinguishes Tumor vs. Normal Benign.

\textbf{iv). DigestPath} \citep{da2022digestpath}: is a dataset of H\&E-stained colonoscopy tissue sections, comprising 660 whole-slide images. Following the protocol used in PLIP, we conducted patch-level zero-shot classification (Tumor vs. Normal) on the official testing set, which includes 18,814 image patches.

\textbf{v). TCGA-BRCA} \citep{tomczak2015review}: is a whole-slide image (WSI) dataset of invasive breast carcinoma derived from The Cancer Genome Atlas (TCGA), comprising two subtypes: Invasive Ductal Carcinoma (IDC) and Invasive Lobular Carcinoma (ILC). It includes a total of 1,048 WSIs, with 837 IDC and 211 ILC slides. For zero-shot WSI-level classification, following the protocol in MI-Zero, we used a test set consisting of 75 WSIs from each class, ensuring no patient-level overlap between training and testing splits.

\textbf{vi). TCGA-RCC} \citep{tomczak2015review}: is a renal cell carcinoma WSI dataset from TCGA consisting of three subtypes: Clear Cell RCC (CCRCC, 519 slides), Papillary RCC (PRCC, 294 slides), and Chromophobe RCC (CHRCC, 109 slides), totaling 922 WSIs. For zero-shot classification, we use 75 WSIs from each subtype as the test set, with no patient overlap between training and testing, following the MI-Zero setup.

The first four datasets are at the patch level, focusing on localized morphological patterns, while the latter two are whole-slide level datasets that require global diagnostic reasoning across multiple cancer subtypes.
\section{Implementation Details}
\label{sec:appendix11}
We implemented all models using PyTorch and conducted training on 2 NVIDIA A100 GPUs. Across all vision-language pretraining variants, we used a temperature parameter of 0.02 for the contrastive loss and optimized using AdamW~\cite{loshchilov2017decoupled} with an initial learning rate of $5 \times 10^{-6}$. A cosine decay scheduler was applied throughout training. All models were trained for 50 epochs with a batch size of 256. During fine-tuning, we adopted \textbf{BioClinicalBERT} (with a maximum token length of 512) as the text encoder, and \textbf{PLIP-ViT-B/32-224} as the visual encoder. All images were resized to $224 \times 224$, and standard data augmentations including random cropping, horizontal flipping, and color jittering were applied during training. For zero-shot classification, we used single prompts per class for evaluation, following the standard setup in prior works. Following a unified protocol, we finetune all pathology-focused baselines on the ARCH training split using their \emph{original} objectives and \emph{without} our perturbation-based augmentations, and then evaluate them on PathoHR-Bench and public datasets with a single prompt per class. 
Although PLIP is pretrained on large biomedical corpora, it is not originally finetuned on ARCH; we therefore additionally finetune PLIP on ARCH to reduce distribution shift. 
To preserve a general-purpose reference, CLIP remains a purely zero-shot baseline (no ARCH fine-tuning). 

\section{Zero-shot without ARCH fine-tuning.}
\label{sec:appendixd}
In Table~\ref{tab:zs_wout_arch}, we observe that the performance differences between the settings with and without fine-tuning on ARCH are not very large, especially for PLIP. 
This is likely because PLIP was already pretrained on a sufficiently large number of pathology-specific image--text pairs, so plain fine-tuning on paired data may approach an upper limit for representation learning on coarser-grained tasks such as PanNuke and DigestPath. 
In contrast, our method yields clear improvements on fine-grained subtype classification (CRC100K and UHU) and on PathoHR-Bench even when trained with only limited ARCH data. 

We attribute these gains to the use of carefully designed structured positives and targeted hard or soft negatives that explicitly supervise compositional reasoning over entities, descriptors, and relations, rather than relying solely on generic image-text alignment. 
Taken together, these results indicate that our framework can more effectively capture the inherent logic and hierarchical structure of pathology data beyond what standard fine-tuning can achieve.

\begin{table}[t]
  \centering
  \fontsize{7}{8}\selectfont
  \setlength{\tabcolsep}{3pt}
  \begin{tabular}{
    >{\centering\arraybackslash}m{2.1cm}|
    >{\centering\arraybackslash}m{0.65cm}
    >{\centering\arraybackslash}m{0.65cm}|
    >{\centering\arraybackslash}m{0.65cm}
    >{\centering\arraybackslash}m{0.65cm}|
    >{\centering\arraybackslash}m{0.65cm}
    >{\centering\arraybackslash}m{0.65cm}
  }
    \toprule
    \multirow{2}{*}{\makecell{\textbf{Model}\\\textbf{(no ARCH finetune)}}} &
    \multicolumn{2}{c|}{\textbf{UHU}} &
    \multicolumn{2}{c|}{\textbf{PanNuke}} &
    \multicolumn{2}{c}{\textbf{DigestPath}} \\
    \cmidrule(lr){2-7}
     & \textbf{Acc} & \textbf{F1} & \textbf{Acc} & \textbf{F1} & \textbf{Acc} & \textbf{F1} \\
    \midrule
    \makecell{BiomedCLIP\\(w/o finetune)} & 0.3371 & 0.1804 & 0.5019 & 0.5073 & 0.5481 & 0.5226 \\
    \makecell{PLIP\\(w/o finetune)}       & 0.3618 & 0.2085 & 0.6139 & 0.6014 & 0.7890 & 0.8027 \\
    \bottomrule
  \end{tabular}
  \caption{\label{tab:zs_wout_arch}
  Zero-shot results \emph{without} ARCH fine-tuning for two widely used baselines. Numbers are reported with a single prompt per class under our unified protocol.}
\end{table}

\section{Case Study}
\label{sec:appendix2}
In this section, we present additional case studies to further demonstrate the improvement in fine-grained learning capabilities achieved by our proposed training scheme. Figure~\ref{fig5} illustrates examples of semantic drift perturbations at the connections level, information loss perturbations at the descriptors level, and order variation perturbations at the entities level within the PathoHR-Bench. Additionally, comparative results are presented on PLIP and our proposed training framework across these tasks, highlighting their performances in handling different types of perturbations. The results demonstrate that our proposed training scheme significantly enhances compositional reasoning and structural awareness, enabling the model to better understand pathological diagnostic logic and reasoning processes.

To further highlight the advantages of our scheme in real-world diagnostic applications, we take a practical diagnostic task as an example. Prostate Gleason grading is a standard pathology assessment used to evaluate the aggressiveness of prostate cancer, which is classified into four grades (Benign, Grade 3, Grade 4 and Grade 5). Although existing models can effectively distinguish between low-grade and high-grade cancers, they often struggle to differentiate between Grade 3 and Grade 4, where the differences are subtle and accurate classification requires attention to fine-grained pathological features, leading to frequent mis-classifications. Such errors can significantly affect prognosis and treatment decisions. 
As shown in the figure~\ref{fig6}, our model exhibits a notable advantage on these challenging borderline cases, demonstrating its superior capability in capturing subtle pathological differences and providing more reliable diagnostic predictions.
\begin{figure*}[htbp]
  \centering
  \includegraphics[width=\linewidth]{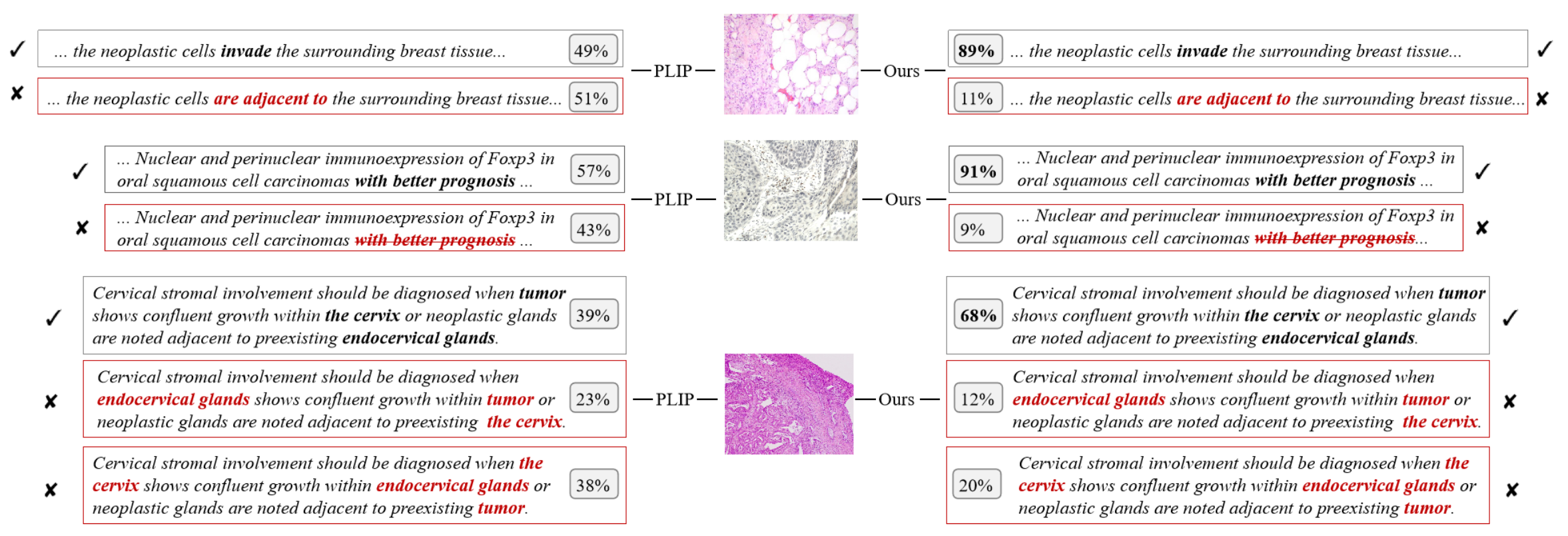}
  \caption {Cases study of semantic drift perturbations at the connections level, information loss perturbations at the descriptors level, and order variation perturbations at the entities level within the PathoHR-Bench.}
  \label{fig5}
\end{figure*}
\begin{figure*}[htbp]
  \centering
  \includegraphics[width=\linewidth]{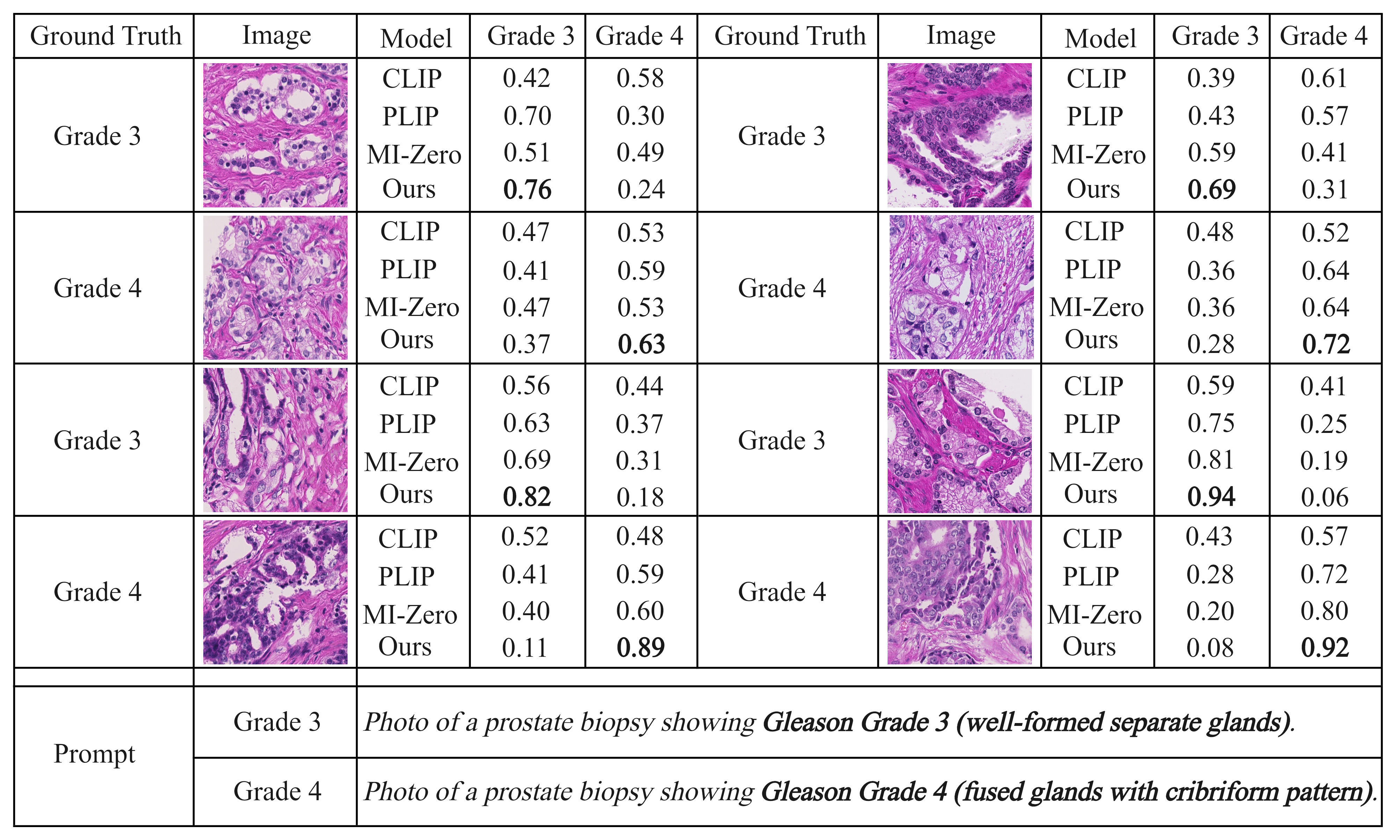}
  \caption {Case Study of fine-grained classification for Gleason grades.}
  \label{fig6}
\end{figure*}

\section{Pathological Relationships in Negative Sample Generation}
\label{sec:appendix3}
In this section, we analyze three types of relationships that may arise when generating negative text samples as shown in Table~\ref{5}. Contrasting relationships refer to pairs of pathological concepts with opposite pathological characteristics or diagnostic properties, while parallel relationships represent concepts at the same hierarchical level but belonging to different pathological categories. Negative samples generated through these two relationships enhance the model’s fine-grained classification capability and category boundary awareness. However, there exists an inclusive relationship, where one pathological concept is a subset of a broader concept, which can result in high semantic similarity between negative samples and the original samples, thereby reducing the effectiveness of contrastive learning.
\begin{table*}[htbp]
  \centering
  \begin{tabular}{cc}
  \hline
  \textbf{Relationship} & \textbf{Case} \\
    \hline
    \textbf{Contrasting Relationship} & in colon \textbf{carcinoma} $\rightarrow$ in colon \textbf{adenoma}\\
    \textbf{Parallel Relationship} & in \textbf{colon} carcinoma $\rightarrow$ in \textbf{gastric} carcinoma\\
    \textbf{Inclusion Relationship} & in \textbf{colon} carcinoma $\rightarrow$ in \textbf{gastrointestinal} carcinoma \\
    \hline
  \end{tabular}
  \caption{\label{5} Pathological relationship types in negative sample generation. 
  }
\end{table*}
\section{Expert Evaluation of Generated Samples}
\label{sec:appendix4}
To assess the clinical relevance and diagnostic validity of the generated samples, we conducted a qualitative review with an expert pathologist.
\subsection{Evaluation Setup}
We randomly selected 200 samples, divided into five categories (40 per type):

\noindent \textbf{i). Text Neg:} generated via semantic perturbation.

\noindent \textbf{ii). Img Neg$_\text{Easy}$:} generated by Stable Diffusion guided by corrupted diagnostic texts.

\noindent \textbf{iii). Img Neg$_\text{Hard}$:} generated through adversarial distribution-aware perturbation.

\noindent \textbf{iv). Text Pos:} generated through hierarchical diagnostic reasoning.

\noindent \textbf{v). Img Pos:} generated via wavelet-morphology-guided consistency refinement.

Each sample was rated by the expert using a 1–5 Likert scale on dimensions relevant to its intended purpose:
\begin{itemize}
\item For positive samples and hard negative images: clinical realism and structural integrity.
\item For text-guided negatives and textual negatives: semantic inconsistency clarity and misleading plausibility.
\end{itemize}
Free-form comments were also collected.
\subsection{Evaluation Summary}
\noindent \textbf{Textual Negatives:} 90\% were judged as plausibly misleading, effectively simulating clinical contradictions. Inclusion-type perturbations were particularly subtle and challenging.

\noindent \textbf{Text-guided Image Negatives:} 85\% of samples exaggerated incorrect structures in a way that was visually interpretable. While not clinically realistic by design, the expert confirmed their usefulness for training models to reject structurally invalid patterns.

\noindent \textbf{Adversarial Image Negatives:} 87.5\% retained realistic tissue appearance while introducing subtle, distribution-aware shifts. Many resembled borderline cases in real practice.

\noindent \textbf{Textual Positives:} 80\% were rated clinically coherent, capturing diagnostic logic. Some samples were considered too generic or lacking critical context for high-grade interpretations.

\noindent \textbf{Image Positives:} 92.5\% preserved diagnostic structures and enhanced local detail (e.g., nuclei, fibrous margins) without introducing artifacts.
\subsection{Expert-Identified Limitations}
During the review, the expert also identified several limitations in the generated samples. Some text-guided image negatives exhibited subtle structural artifacts or biologically implausible tissue combinations. In several cases, semantic perturbations were considered too trivial to mislead a diagnostic system. For positive text expansions, certain samples lacked sufficient diagnostic context, such as staging information or relevant biomarkers. These observations provide valuable feedback for refining the sample generation strategies and improving their clinical fidelity.
\end{document}